\documentclass[letterpaper, 10 pt, conference]{ieeeconf}  

\IEEEoverridecommandlockouts                              

\usepackage{times}
\usepackage{epsfig}
\usepackage{graphicx}
\usepackage{amsmath}
\usepackage{amssymb}
\usepackage{times}
\usepackage{soul}
\usepackage{url}
\usepackage{subcaption}
\usepackage{booktabs}
\usepackage{float}
\usepackage{algorithm}
\usepackage{algpseudocode}

\usepackage[breaklinks=true,bookmarks=false]{hyperref}
\title{\LARGE \bf
Test-Time Adaptation for Point Cloud Upsampling Using Meta-Learning
}

\author{{Ahmed Hatem$^{1}$, Yiming Qian$^{2}$, Yang Wang$^{3}$}
\thanks{$^{1}$University of Manitoba.
        {\tt\small hatema@myumanitoba.ca}}%
\thanks{$^{2}$University of Manitoba.
        {\tt\small yiming.qian@umanitoba.ca}}%
\thanks{$^{3}$Concordia University.
        {\tt\small yang.wang@concordia.ca}}%
}

\begin{document}

\maketitle
\thispagestyle{empty}
\pagestyle{empty}

\begin{abstract}
Affordable 3D scanners often produce sparse and non-uniform point clouds that negatively impact downstream applications in robotic systems. While existing point cloud upsampling architectures have demonstrated promising results on standard benchmarks, they tend to experience significant performance drops when the test data have different distributions from the training data. To address this issue, this paper proposes a test-time adaption approach to enhance model generality of point cloud upsampling. The proposed approach leverages meta-learning to explicitly learn network parameters for test-time adaption. Our method does not require any prior information about the test data. During meta-training, the model parameters are learned from a collection of instance-level tasks, each of which consists of a sparse-dense pair of point clouds from the training data. During meta-testing, the trained model is fine-tuned with a few gradient updates to produce a unique set of network parameters for each test instance. The updated model is then used for the final prediction. Our framework is generic and can be applied in a plug-and-play manner with existing backbone networks in point cloud upsampling. Extensive experiments demonstrate that our approach improves the performance of state-of-the-art models.
\end{abstract}
\section{Introduction}
Point cloud is one of the most popular representations of the 3D information in robotic applications. Point clouds can be obtained from readily available 3D scanning devices, such as LiDARs and RGB-D cameras. They play a vital role in a variety of applications, such as autonomous driving \cite{3Dauto}, augmented reality \cite{ar_survey}, and robotics \cite{robotics_survey}. However, the point clouds obtained from affordable 3D scanners are usually sparse and non-uniform. Therefore, these sparse point clouds need to be effectively upsampled to produce denser point clouds in order to be used in many downstream applications. 

\par Given an input sparse point cloud, our goal is to generate a high-resolution uniform point cloud that adequately represents the underlying surface. Many traditional optimization-based methods \cite{pu_optim1,pu_optim2,pu_optim3,pu_optim4,pu_optim5} have been proposed for point cloud upsampling. Recently, DNN-based upsampling methods have emerged and achieved impressive results, including MPU \cite{mpu}, PUGAN \cite{pugan}, PUGCN \cite{pugcn}, Dis-PU \cite{dispu}, and PU-Dense \cite{pudense}. These methods can learn complex structures of point clouds and outperform traditional approaches. However, most learning-based methods have followed a fully supervised learning paradigm for point cloud upsampling. They assume that the training and test data are sampled from the same data distribution. This is unrealistic for real-world scenarios due to the fact that the data captured by different 3D sensors have large discrepancies. It is challenging for the training data to cover all the variations that can happen  during testing. Consequently, trained models usually experience a drastic drop in performance when they are evaluated on unknown test distributions. This is known as the domain shift problem.

\par To address the distribution shift issue and the resulting performance drop, recent work has proposed domain adaptation approaches for point clouds to minimize the gap between training and test distributions, e.g. using adversarial learning \cite{pointdan,geoAdapt,advAdapt,segmentAdapt} or self-supervised learning \cite{sslAdapt,sslAdapt3,sslAdapt4,sslAdapt5}. However, these approaches have some limitations. First, they assume having access to unlabeled samples of the target test distribution during training. This may not be feasible in real-world settings where the information from test distribution is not available in advance. In addition, they do not fully utilize the useful internal information available within the test instance, since the trained model parameters are fixed during inference time for all unseen test instances.

\par Recently, Zhou et al. \cite{zspu} have adopted ZSSR \cite{zssr} for point cloud upsampling. They have proposed a Zero-Shot Point Cloud Upsampling (ZSPU) \cite{zspu} approach that can capture the internal information provided by a given point cloud at test time. This method trains the network from scratch at test time using augmented pairs of sparse-dense point clouds extracted from the test point cloud. Although ZSPU \cite {zspu} can successfully exploit the internal features of the test instances, it requires a high inference time due to self-training at test time. In addition, it fails to utilize the useful information learned from the external dataset.

\par In this work, we address the above limitations by introducing a test-time adaptation approach that leverages the internal and external information of point clouds. Concretely, we first conduct large-scale training using pairs of sparse-dense point clouds to utilize the external dataset. Then, we adapt the model parameters in an instance-specific manner during inference and obtain a different set of network parameters for each different instance. This allows our model to better capture the uniqueness of each test sample and thus generalize better to unseen data. We have found that a simple fine-tuning of the pre-trained network is not optimal, since it  takes a huge number of gradient updates for adaptation. Therefore, we propose to use meta-learning for a fast and effective adaptation of the model at test time. Meta-learning has shown great success in learning new tasks quickly with few training samples. In particular, Model-Agnostic Meta-Learning (MAML) \cite{maml} has been widely employed for various domain adaptation tasks \cite{metaResolution,metaVideo,deblurring_2021_cvpr,maxl}. MAML \cite{maml} is an optimization-based method that aims to learn the model parameters in a way that facilitates fast adaptation at test time within a few gradient updates. We adopt MAML \cite{maml} for training the point cloud upsampling networks. During meta-training, each input point cloud is downsampled by a predefined scaling factor and the MAML \cite{maml} task is the reconstruction of the input point cloud. At test time, the model is updated by a few gradient updates based on a self-supervised learning procedure that exploits the internal information of the test point cloud. The updated model is then used for the final prediction. Our key contributions are summarized as follows:
\begin{itemize}
\item We propose a test-time adaptation approach for point cloud upsampling. To the best of our knowledge, this is the first work that exploits the complementary advantages of both internal and external learning for point cloud upsampling.
\item We propose to employ meta-learning to provide the model with the ability of fast and effective adaptation at inference time to improve the model generalization. 
\item We introduce a novel model-agnostic framework that can be applied on   any point cloud upsampling network to boost its performance. 
\end{itemize}
\section{Related Work}
\newcommand{\rulesep}{\unskip\ }
\begin{figure*}
\begin{center}
\begin{subfigure}[b]{0.64\textwidth}
\includegraphics[width=\textwidth]{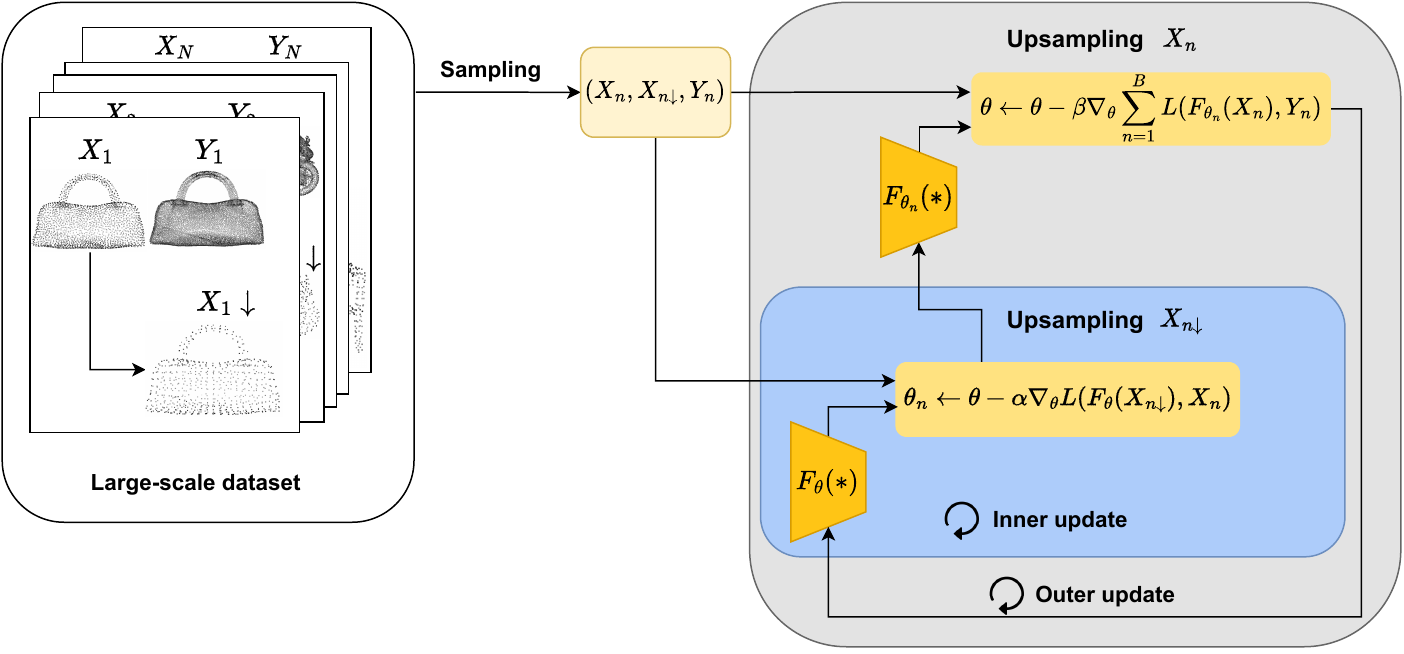}
\caption{Meta-training}
\end{subfigure}
\hfill
\begin{subfigure}[b]{0.34\textwidth}
\includegraphics[width=\textwidth]{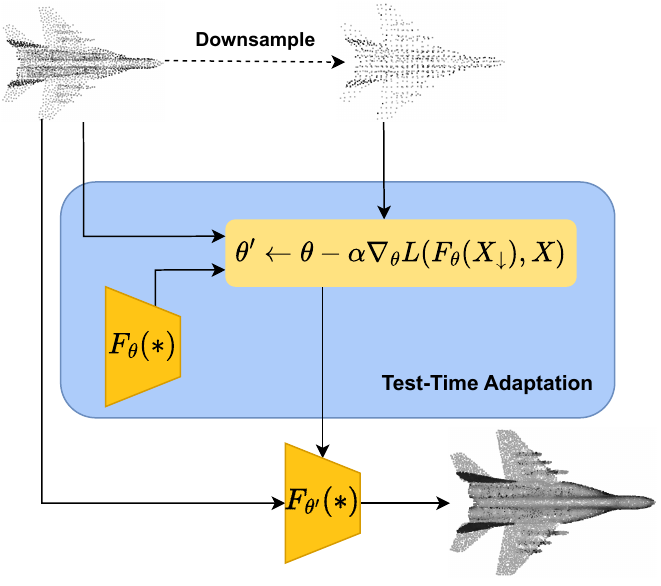}
\caption{Meta-testing}
\end{subfigure}
\end{center}
   \caption{Overview of the proposed meta-learning procedure for point cloud upsampling. During each iteration of meta-training, we sample a batch of training pairs. For each sampled training pair $(X_n,Y_n)$, we first downsample $X_n$ to obtain a sparser version $X_{n\downarrow}$. We obtain the adapted parameters by applying our model to upsample $X_{n\downarrow}$ and use $X_n$ as the ground-truth to define as a self-supervised loss. We perform a small number of gradient updates using the computed self-supervised loss in the inner loop. Then we use the adapted model to perform the main task by upsampling $X_{n}$. Finally, we update the model in the outer loop based on the calculated upsampling loss on the adapted parameters. Given a new instance $X$ during meta-testing, we perform a few gradient updates using $(X_{\downarrow},X)$ to adapt the model for this instance and use the adapted model for final prediction.}
\label{fig:overview}
\end{figure*}

Our work is closely related to several lines of research. We briefly review prior work closest related to ours.
\subsection{Point Cloud Upsampling}
Early work has proposed optimization-based methods  \cite{pu_optim1,pu_optim2,pu_optim3,pu_optim4} for densifying point clouds. Alexa et al. \cite{pu_optim1} generate dense points by inserting points at the Voronoi diagram’s vertices. Lipman et al. \cite{pu_optim2} introduce a locally optimal projection operator for resampling point clouds based on L1 norm. Later, an improved weighted version of the local optimal projection operator was proposed in \cite{pu_optim3}. However, these methods usually do not work well around sharp edges. Huang et al. \cite{pu_optim4} propose an edge-aware resampling algorithm that progressively transfers generated points towards the edge singularities to preserve edge sharpness. 

\par Ever since the emergence of deep learning, recent work has shifted to upsampling  point clouds using deep neural networks. PU-Net \cite{punet} is the first to apply deep learning in point cloud upsampling. It uses Point-Net++ \cite{pointnetplus} for feature extraction and expands point features using multi-branch convolutions in feature space. EC-Net \cite{puec} proposes an edge-aware point cloud upsampling by directly minimizing distances from points to edges. MPU \cite{mpu} proposes a patch-based network that learns different levels of point cloud features by progressively upsampling the points in multiple steps. PU-GAN \cite{pugan} adopts a generative adversarial network to generate high-quality upsampled points. PU-GCN \cite{pugcn} proposes a graph convolutional network for point cloud upsampling. PUGeo-Net \cite{pugeo} incorporates differential geometry to improve point cloud upsampling performance by learning the local geometry of point clouds. Dis-PU \cite{dispu} introduces disentangled refinement units for point upsampling using two sub-networks, including a dense point generator and a point spatial refiner. PU-Dense \cite{pudense} proposes a novel feature extraction unit for extracting 3D multiscale features and adopts U-Net architecture based on sparse convolutions for computationally efficient processing of point clouds. 

\subsection{Domain Adaptation}
Extensive work has been proposed for 2D image domain adaptation \cite{imageAdapt2,imageAdapt3,imageAdapt4,imageAdapt5}. Recently, there is also work exploring domain adaptation for point clouds. Most existing domain adaptation approaches on point clouds \cite {pointdan,rangeAdapt,bevAdapt,geoAdapt} mainly rely on adversarial learning to transfer knowledge from labeled source domain to unlabeled target domain. Qin et al. introduce PointDAN \cite{pointdan} to jointly align local and global features of point cloud distributions across different domains for 3D classification. Wang et al. \cite{rangeAdapt} propose a cross-range adaptation to enhance far-range 3D object detection performance. Saleh et al. \cite{bevAdapt} adopt CycleGAN \cite{cycleGAN} to adapt projected 2D bird’s eye view synthetic images for real-world vehicle detection. Wu et al. \cite{geoAdapt} use geodesic correlation alignment for minimizing the domain gap between synthetic and real data. 
\par Recently, several studies have proposed to design self-supervised tasks \cite{sslAdapt1,sslAdapt2,sslAdapt3,sslAdapt4} for learning domain invariant features of point clouds. \cite{sslAdapt1} introduces a deformation reconstruction task to learn the underlying structures of 3D objects. \cite{sslAdapt2} defines the self-supervised task as a reconstruction of random partially displaced point clouds. \cite{sslAdapt3} proposes two self-supervised tasks, including a scale prediction task and a 3D/2D projection reconstruction task to transfer global and local features across domains. 
\par The main limitation of most existing domain adaptation approaches is the assumption of the availability of unlabelled target domain data during training, which is infeasible when the target domain is unknown during training.

\subsection{Meta Learning}
\par Meta-learning has been successfully applied in many computer vision and robotics applications. Existing meta-learning methods can be categorized into
model-based \cite{metaModel1,metaModel2,metaModel3,metapu,meta3Dseg}, metric-based \cite{metaMetric1,metaMetric2,metaMetric3,metaMetric4} and optimization-based methods \cite{maml,metaOptim1,metaOptim2,metaOptim3}. Model-based methods learn to update their model parameters with a few steps either using another meta-learner network for parameters prediction \cite{metapu,meta3Dseg} or via its internal architecture \cite{metaModel1}. Metric-based methods learn a metric function to measure the similarity between samples. Optimization-based methods learn an optimal model initialization that can rapidly adapt to new tasks. 
\par MAML \cite{maml} is a widely used optimization-based algorithm, which has been successfully adapted to many 2D image domain tasks \cite{metaGAN,deblurring_2021_cvpr,meta_depth,metaResolution,metaResolution2}. Zhang et al. \cite{metaGAN} propose MetaGaN that integrates the MAML algorithm with GAN network to improve image classification performance with a few number of training samples. Chi et al. \cite{deblurring_2021_cvpr} introduce meta-auxiliary learning framework based on MAML for image deblurring to enable fast model adaptation. Liu et al. \cite{meta_depth} propose to use meta-auxiliary learning with test-time adaptation for the problem of future depth prediction in videos. \cite{metaResolution,metaResolution2} use MAML for image super-resolution, which learns effective pre-trained model weights that can quickly adapt to unseen test images. Relatively little meta-learning work has been proposed for point clouds \cite{metapu,meta3Dseg,metasets}. Hai et al. \cite{meta3Dseg} introduce a parameter prediction network for point cloud segmentation to enable fast adaptation to new part segmentation tasks. Ye et al. \cite{metapu} propose a meta-subnetwork for point cloud upsampling that is trained to dynamically adjust the upsampling network parameters to support flexible scale factors.
\section{Proposed Method}

Given a sparse and noisy point cloud $ X \in R^{N \times 3}$ with $N$ points and an upsampling ratio $r$, our goal is to generate a dense point cloud $Y \in  R^{ rN \times 3}$ with $rN$ points that adequately cover the underlying surface. More importantly, the generated points should be uniformly-located on the object surface. We aim to learn a model $F_{\theta}(X)\rightarrow Y$ parameterized by $\theta$ that maps $X$ to $Y$ for a given
upsampling ratio $r$. 

\subsection{Preliminaries}
To leverage the advantages of external learning, we first perform supervised training using pairs of sparse-dense point clouds $(X,Y)$. We adopt the architecture of three state-of-the-art point cloud upsampling networks as the backbones of our approach, including PU-GCN \cite{pugcn}, Dis-PU \cite{dispu}, and PU-Dense \cite{pudense}. Each network is optimized using a standard supervised loss to learn the initial model parameters $\theta$:
\begin{equation}
\label{eq:supervise}
\min_{\theta} L(F_{\theta}(X),Y)
\end{equation}
where $L$ is a supervised loss between the prediction $F_{\theta}(X)$ and the ground truth $Y$.
\par At this step, we may directly fine-tune the pre-trained parameters during inference to exploit the internal features of point clouds for test-time adaptation (TTA). For example, given an input point cloud $X$ during inference, we can downsample $X$ to obtain a sparser point cloud $X_{\downarrow}$. We can then fine-tune the model parameter $\theta$ by treating $(X_{\downarrow}, X)$ as a sparse-dense pair of point clouds in Eq~\ref{eq:supervise}. In our experiments, we will demonstrate that such naive TTA can already improve the model performance. However, it requires a large number of gradient updates to effectively adapt for each test instance since the model is not explicitly learned to facilitate test-time adaptation. 

In our work, we propose a meta-learning approach to explicitly learn the model parameters for test-time adaptation (Fig.~\ref{fig:overview}). Our approach consists of a meta-training stage and a meta-testing stage. During meta-training, we learn the model parameters from a set of tasks, where each task is constructed from a sparse-dense pair of point clouds from the training data. Each inner update of meta-training involves adapting the model to a sampled task. The adapted model is then used for inference on that task. This performance of this task-adaptive model is then used for the loss function for the outer update of meta-training. The goal of the outer update is to optimize the model parameters such that after adapting to a particular task, the adapted model performs well on that task. Through this bi-level optimization, the model parameters are explicitly trained so that they can be effectively adapted to a new test instance with only a few gradient updates. During meta-testing, we are given a new test instance. We first adapt the meta-learned model to this instance, then use the updated model for prediction.

\subsection{Meta-Training}
\par Inspired by the success of adopting MAML \cite{maml} for image super-resolution problem \cite{metaResolution,metaResolution2}, we learn the model using meta-learning such that the model parameters are trained to quickly adapt to unseen data at test time using a small number of gradient updates. First, the model is initialized by the pre-trained weights $\theta$ resulting from the standard supervised training. The pre-trained feature representations help in stabilizing meta-training and thus ease the training phase of meta-learning  \cite{metaResolution2}. We further optimize the model parameters using meta-learning. Specifically, we develop a meta-learning algorithm summarized in Algorithm \ref{alg:algoTrain} based on MAML \cite{maml}. The key to our approach is the construction of a task for the inner update of meta-training. We use a pair of point clouds consisting of the input point cloud $X$ and its downsampled version $X_{\downarrow}$ for a task in MAML. During the inner update, the model is used to upsample $X_{\downarrow}$ and $X$ is treated as the ground truth. The loss between $F_{\theta}(X_{\downarrow})$ and $X$ is  then used to adapt the model parameters $\theta$ by a few gradient updates. This allows the model to be quickly adapted to different data distributions at test time and boosts the overall generalization capability of the model. 
\par Figure \ref{fig:overview} illustrates the overall scheme of our proposed approach. The external training dataset consists of pairs of sparse-dense point clouds. We optimize the network weights using our proposed meta-learning approach to learn the optimal model parameters that can quickly adapt to new data distributions at test time. At test-time, we adapt the meta-learned parameters for each given test instance and use the adapted parameters to obtain the upsampled point cloud $Y$.
\par More specifically, in each iteration of meta-training, we sample a batch $B$ of sparse-dense training pairs $\{X_{n}, Y_{n}\}_{n=1}^B$. We downsample $X_{n}$ to a sparser version $X_{n\downarrow}$. In each inner update of meta-training, we perform model adaptation for a small number of gradient updates using ($X_{n\downarrow}$,$X_{n}$) pairs as follows:
\begin{equation}
\theta_{n} \leftarrow \theta - \alpha \nabla_{\theta} L(F_{\theta}(X_{n\downarrow}),X_{n})
\end{equation}
where $\alpha$ controls the learning rate of the adaptation and $\theta_n$ represent the adapted model parameters for the input $X_n$ using internal learning.

The adapted model $\theta_n$ is then used to generate the dense point cloud $Y$ and optimize the following meta-objective:
\begin{equation}
\label{eq:metaobj}
\min_{\theta} \sum_{n=1}^B L(F_{\theta_n}(X_n),Y_n)
\end{equation}
Note that we use the adapted model $\theta_{n}$ in the model $F_{\theta_{n}(\cdot)}$, but the optimization in Eq.~\ref{eq:metaobj} is performed over the original model parameters $\theta$.

In the outer update of meta-training, we optimize the meta-objective in Eq.~\ref{eq:metaobj} by performing gradient update as follows:
\begin{equation}
\theta \leftarrow \theta - \beta \nabla_{\theta} \sum_{n=1}^B L(F_{\theta_n}(X_n),Y_n)
\end{equation}
where $\beta$ is the meta-learning rate.

\subsection{Meta-Testing}
At test-time, we downsample the input point cloud $X$ to a sparser version $X_{\downarrow}$. Then, we fine-tune the model parameters by performing a small number of gradient updates using the point cloud pairs $(X_{\downarrow},X)$. This update is completely self-supervised and exploits the internal features of the input point cloud $X$. 
\begin{equation}
\theta' \leftarrow \theta - \alpha \nabla_{\theta} L(F_{\theta}(X_{\downarrow}),X)
\end{equation}

Finally, the adapted model $\theta'$ is used to perform the main upsampling task and generates the densified point cloud $F_{\theta'}(X)$. The inference procedure is summarized in Algorithm \ref{alg:algoTest}. 

\begin{algorithm}[tb]
\caption{Meta-training}
\label{alg:algoTrain}
\textbf{Require}: $X,Y$: training pairs\\
\textbf{Require}: $B$: batch size\\
\textbf{Require}: $\alpha,\beta$: learning rates \\
\textbf{Output}: $\theta$: learned parameters
\begin{algorithmic}[1] 
\State Initialize the network with pre-trained weights $\theta$
\While{\emph{not converged}}
\State Sample a training batch $\{X_{n},Y_{n}\}^B_{n=1}$
\State Generate downsampled $X_{n\downarrow}$
\For{\emph{n}=1 \emph{to} B}
\State Evaluate loss: $\nabla_{\theta} L(F_{\theta}(X_{n\downarrow}),X_{n})$
\State Compute adapted parameters $\theta_{n}$:
\State $\theta_{n} \leftarrow \theta - \alpha \nabla_{\theta} L(F_{\theta}(X_{n\downarrow}),X_{n})$
\EndFor
\State Evaluate the main upsampling task using the adapted parameters and update:
\State $\theta \leftarrow \theta - \beta \nabla_{\theta} \sum_{n=1}^B L(F_{\theta_n}(X_n),Y_n)$
\EndWhile   
\State \textbf{return} $\theta$
\end{algorithmic}
\end{algorithm}

\begin{algorithm}[tb]
\caption{Meta-testing}
\label{alg:algoTest}
\textbf{Require}: $X$: sparse point cloud\\
\textbf{Require}: $\alpha$: learning rate \\
\textbf{Output}: $Y$: dense point cloud
\begin{algorithmic}[1] 
\State Initialize the network with meta-trained weights $\theta$
\State Generate downsampled $X_{\downarrow}$
\State Evaluate loss: $\nabla_{\theta} L(F_{\theta}(X_{\downarrow}),X)$
\State Compute adapted parameters:
\State $\theta' \leftarrow \theta - \alpha \nabla_{\theta} L(F_{\theta}(X_{\downarrow}),X)$
\State Generate upsampled point cloud Y = $F_{\theta'}(X)$
\State \textbf{return} $Y$
\end{algorithmic}
\end{algorithm}
\section{Experiments}
\begin{figure*}
\centering
      \begin{subfigure}[b]{0.24\textwidth}
         \centering
        \includegraphics[width=0.98\textwidth, height=5cm]{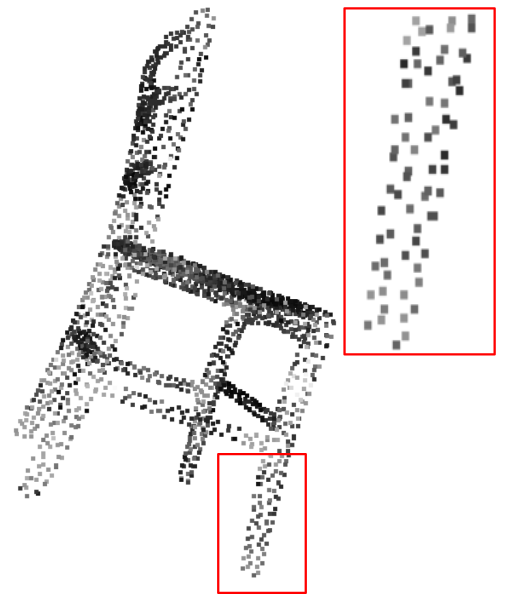}
    \caption*{Input}
     \end{subfigure}%
     \hfill%
      \begin{subfigure}[b]{0.24\textwidth}
         \centering
         \includegraphics[width=0.98\textwidth, height=5cm]{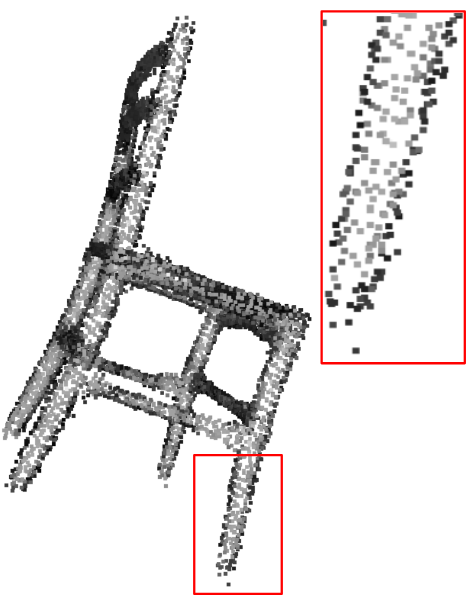}
       \caption*{PU-GCN}
     \end{subfigure}%
     \hfill%
\begin{subfigure}[b]{0.24\textwidth}
         \centering
         \includegraphics[width=0.98\textwidth, height=5cm]{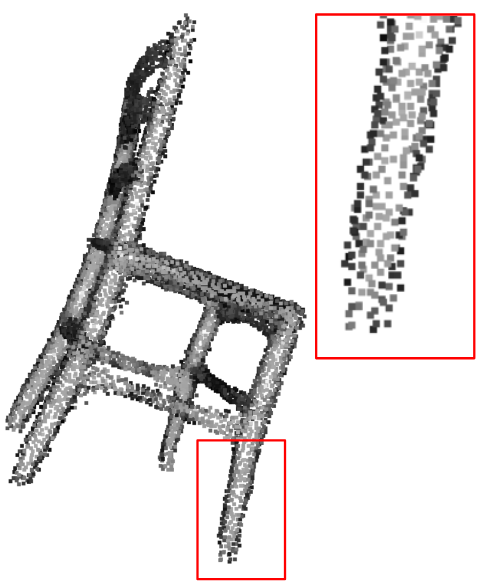}
         \caption*{Ours + PU-GCN}
     \end{subfigure}
    \hfill
    \begin{subfigure}[b]{0.24\textwidth}
         \centering
         \includegraphics[width=0.98\textwidth, height=5cm]{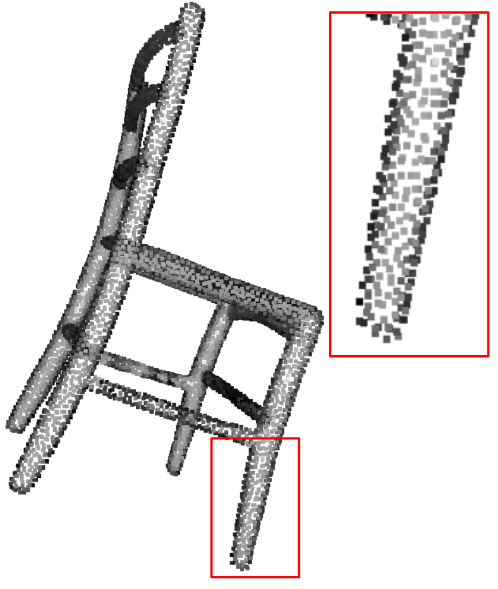}
        \caption*{GT}
     \end{subfigure}%
     \hfill%
     \begin{subfigure}[b]{0.24\textwidth}
         \centering
        \includegraphics[width=0.98\textwidth,height=4cm]{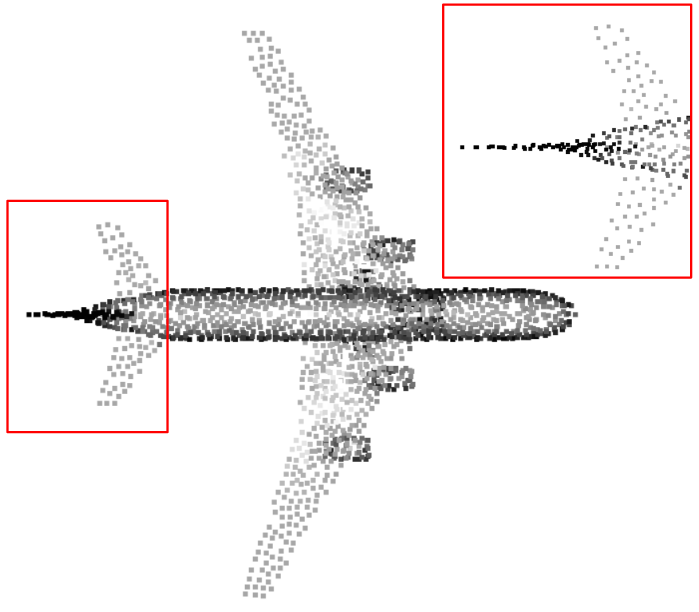}
    \caption*{Input}
     \end{subfigure}%
     \hfill%
      \begin{subfigure}[b]{0.24\textwidth}
         \centering
         \includegraphics[width=0.98\textwidth, height=4cm]{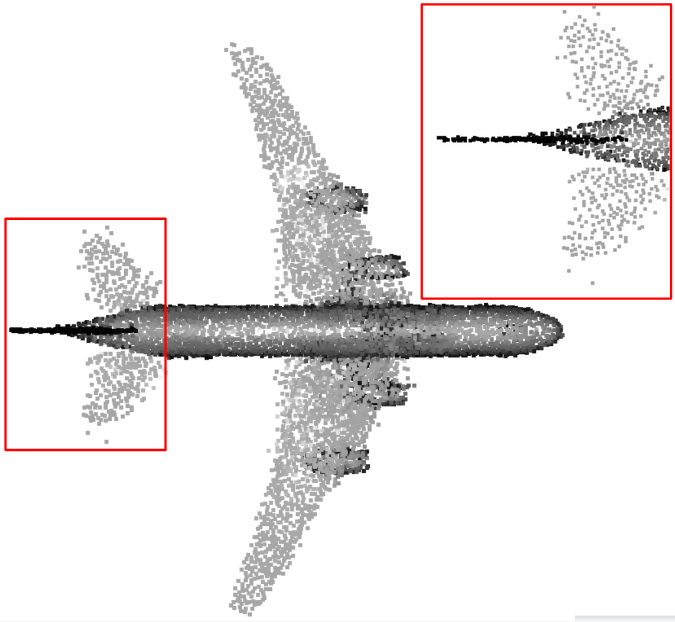}
       \caption*{Dis-PU}
     \end{subfigure}%
     \hfill%
\begin{subfigure}[b]{0.24\textwidth}
         \centering
         \includegraphics[width=0.98\textwidth, height=4cm]{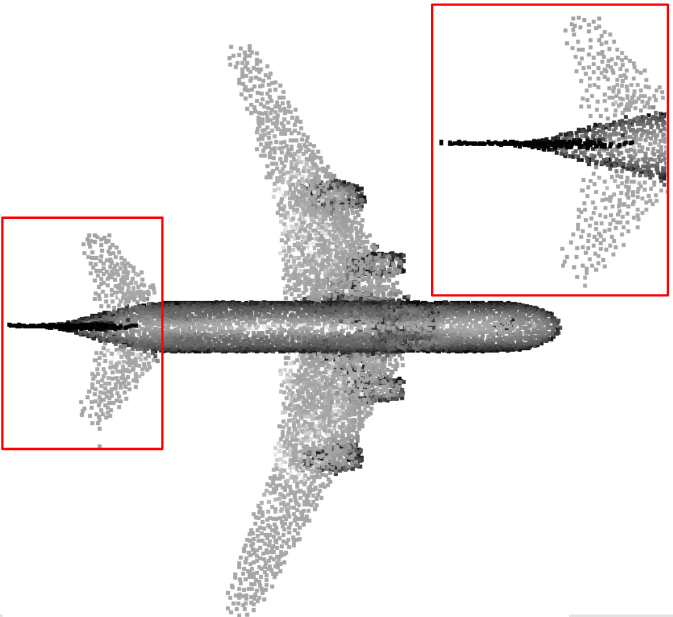}
         \caption*{Ours + Dis-PU}
     \end{subfigure}%
     \hfill%
    \begin{subfigure}[b]{0.24\textwidth}
         \centering
         \includegraphics[width=0.98\textwidth, height=4cm]{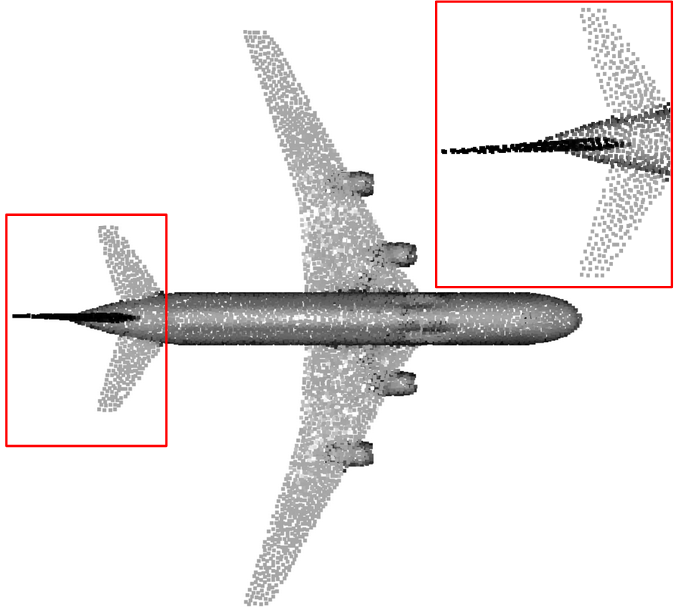}
        \caption*{GT}
     \end{subfigure}%
     \hfill%
     \begin{subfigure}[b]{0.24\textwidth}
         \centering
        \includegraphics[width=0.98\textwidth,height=5cm]{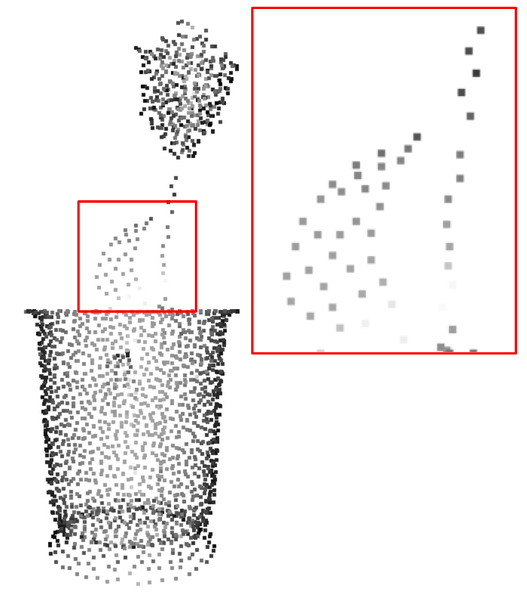}
    \caption*{Input}
     \end{subfigure}%
     \hfill%
      \begin{subfigure}[b]{0.24\textwidth}
         \centering
         \includegraphics[width=0.98\textwidth,height=5cm]{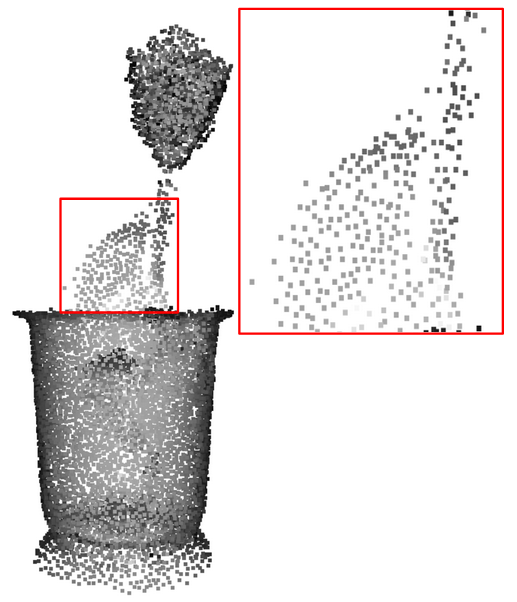}
       \caption*{PU-Dense}
     \end{subfigure}%
     \hfill%
\begin{subfigure}[b]{0.24\textwidth}
         \centering
         \includegraphics[width=0.98\textwidth,height=5cm]{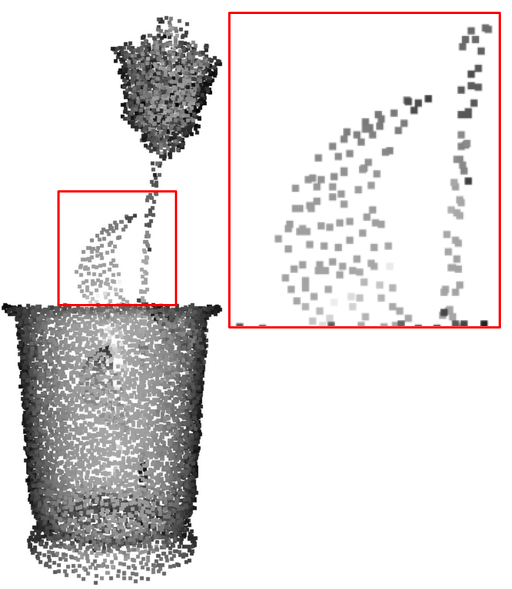}
         \caption*{Ours + PU-Dense}
     \end{subfigure}%
    \hfill%
    \begin{subfigure}[b]{0.24\textwidth}
         \centering
         \includegraphics[width=0.98\textwidth,height=5cm]
         {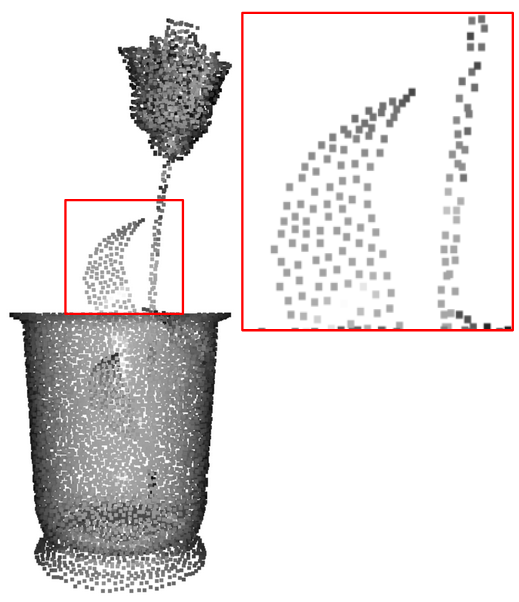}
        \caption*{GT}
     \end{subfigure}%
     \hfill%
   \caption{Qualitative results of point cloud upsampling on the ShapeNet dataset \cite{shapenet}. We compare the 4x upsampled results with the three baselines: PU-GCN \cite{pugcn}, Dis-PU \cite{dispu}, and PU-Dense \cite{pudense}.
}
\vspace{-10pt}

\label{fig:qual-results}
\end{figure*}

\begin{figure*}
\centering
      \begin{subfigure}[b]{0.24\textwidth}
         \centering
        \includegraphics[width =0.98\textwidth, height=5cm]{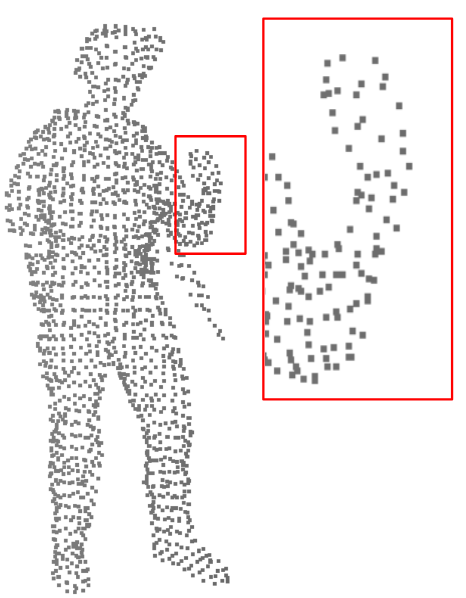}
    \caption*{Input}
     \end{subfigure}
     \hfill
      \begin{subfigure}[b]{0.24\textwidth}
         \centering
         \includegraphics[width =0.98\textwidth, height=5cm]{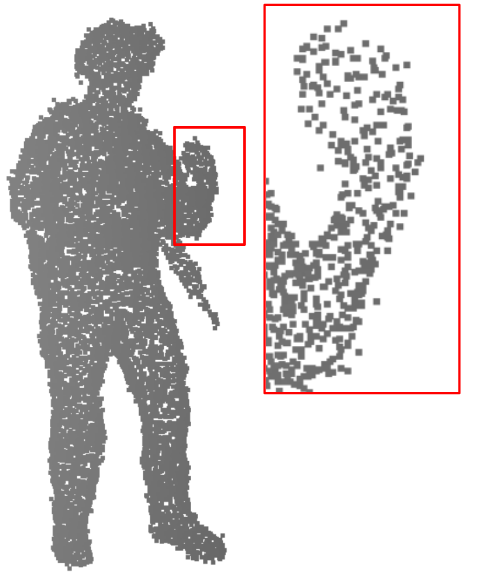}
       \caption*{Dis-PU}
     \end{subfigure}
     \hfill
\begin{subfigure}[b]{0.24\textwidth}
         \centering
         \includegraphics[width =0.98\textwidth, height=5cm]{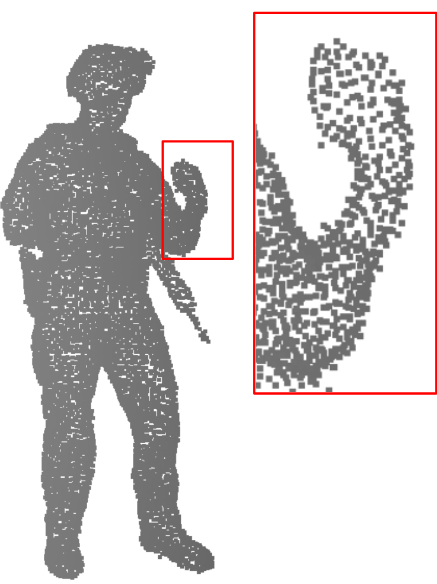}
         \caption*{Ours + Dis-PU}
     \end{subfigure}
    \hfill
    \begin{subfigure}[b]{0.24\textwidth}
         \centering
         \includegraphics[width =0.98\textwidth, height=5cm]{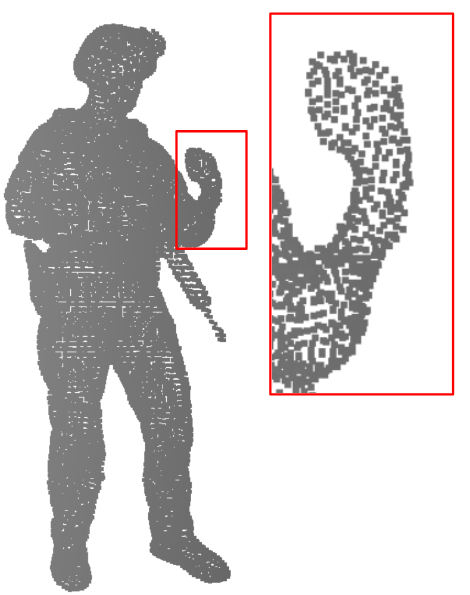}
        \caption*{GT}
     \end{subfigure}
     \hspace{0.5em}

     \begin{subfigure}[b]{0.24\textwidth}
         \centering
        \includegraphics[width =0.98\textwidth, height=5cm]{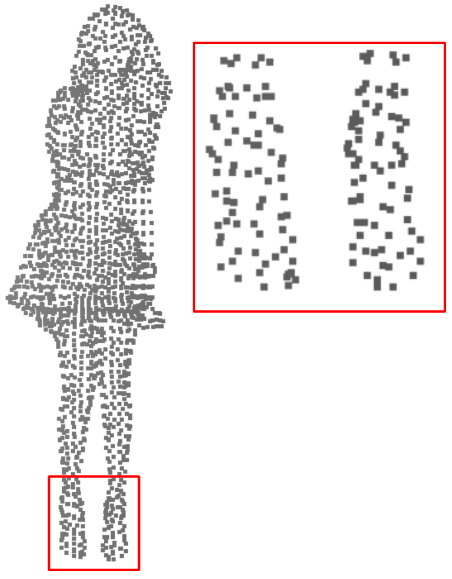}
    \caption*{Input}
     \end{subfigure}
     \hfill
      \begin{subfigure}[b]{0.24\textwidth}
         \centering
         \includegraphics[width =0.98\textwidth, height=5cm]{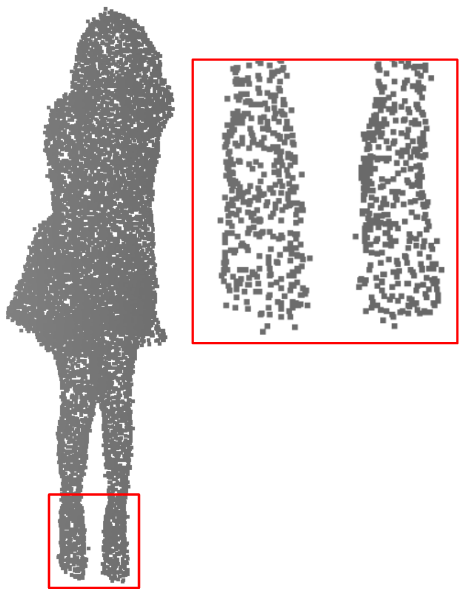}
       \caption*{PU-Dense}
     \end{subfigure}
    \hfill
\begin{subfigure}[b]{0.24\textwidth}
         \centering
         \includegraphics[width =0.98\textwidth, height=5cm]{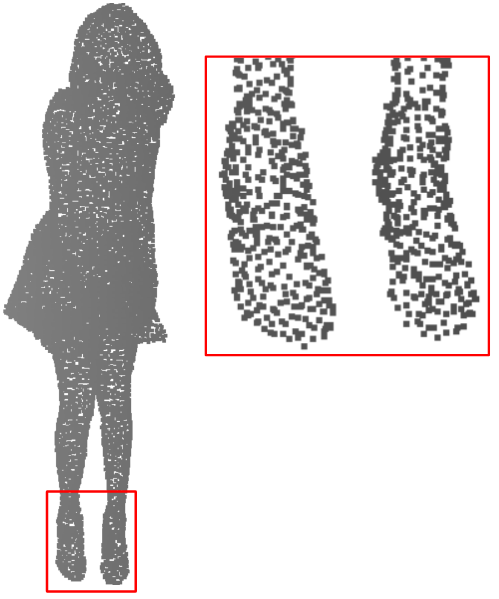}
         \caption*{Ours + PU-Dense}
     \end{subfigure}
    \hfill
    \begin{subfigure}[b]{0.24\textwidth}
         \centering
         \includegraphics[width =0.98\textwidth, height=5cm]{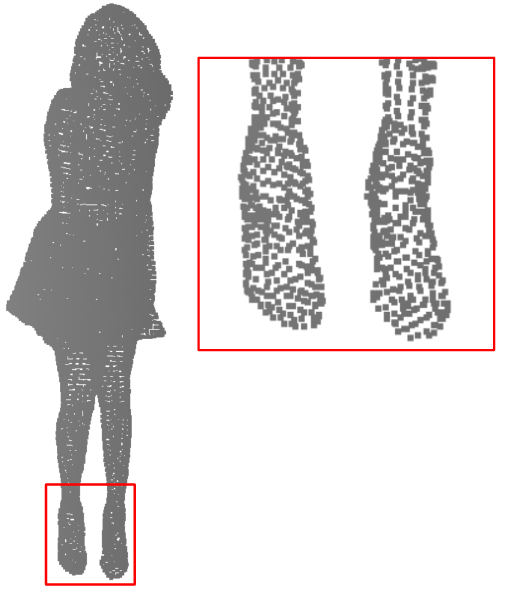}
        \caption*{GT}
     \end{subfigure}
     \hspace{0.5em}
   \caption{Qualitative comparison on the 8iVFB dataset \cite{8i} between the baselines (Dis-PU \cite{dispu} and PU-Dense \cite{pudense}) and our method.}
\label{fig:qual-8i-results}
\end{figure*}

We first describe some implementation details (Sec.~\ref{sec:implementation}). We then introduce the datasets and experimental setup~(Sec.~\ref{sec:setup}). We present our experiment results and comparison in Sec.~\ref{sec:result}. We also perform extensive ablation studies in Sec.~\ref{sec:ablation}.

\subsection{Implementation Details}\label{sec:implementation}

\par We use the official implementations of PU-GCN \cite{pugcn}, Dis-PU \cite{dispu} and PU-Dense \cite{pudense} as the backbones of our approach. We first conduct supervised training on the backbones to obtain the initial pre-trained weights. During meta-training, we perform 5 gradient updates in the inner loop as described in Algorithm \ref{alg:algoTrain}. We set the batch size to 8 and the learning rates $\alpha$ and $\beta$ to $10^{-5}$ and $10^{-6}$, respectively. We optimize the networks using the Adam optimizer with a learning rate of $10^{-4}$ and an exponentially decayed factor of 0.99. All experiments are conducted on a single NVIDIA TitanX GPU.

\subsection{Datasets and Setup}\label{sec:setup}
\par Following \cite{pudense}, we train our method on the ShapeNet dataset \cite{shapenet} and evaluate the performance of the networks on the ShapeNet \cite{shapenet} dataset (1024 test samples), the 8iVFB dataset \cite{8i} (1200 test samples), and the Semantic3D.net dataset \cite{semantic3D} (1500 test samples). We adopt two widely used upsampling evaluation metrics, namely Chamfer distance (CD) and Peak signal-to-noise ratio (PSNR), to measure the quality of the upsampled point cloud compared to the ground truth dense point cloud. CD sums the distances between the nearest neighbors correspondences of the upsampled point cloud  and ground truth point cloud. CD is defined as:
\begin{equation}
 CD_{(Y,G)} = \sum_{a \in Y} \min_{b \in G} |a - b|^2 + \sum_{b \in G} \min_{a \in Y} |a - b|^2
\end{equation}
where $Y$ is the upsampled point cloud and $G$ is the ground truth point cloud.
PSNR measures the ratio between a normalization factor and point-to-point MSE as defined in \cite{pudense}, which is calculated from the upsampled point cloud $Y$ to the ground truth $G$ as well as in the opposite direction. PSNR is defined as follows:
\begin{equation}
    PSNR = min(PSNR_{(Y,G)}, PSNR_{(G,Y)})
\end{equation}
\begin{equation}
\label{eq:psnr}
PSNR_{(A,B)} = 10\log_{10}(\frac{p_s^2}{d_{MSE}(A,B)})
\end{equation}
where $p_s$ is the normalization factor and $d_{MSE}$ is the average mean squared error between points in one point cloud and their nearest neighbors correspondences in the other point cloud.
\subsection{Results and Comparisons}\label{sec:result}
\begin{table}
\caption{Quantitative comparison of our method with existing state-of-the-art methods on ShapeNet \cite{shapenet}, 8iVFB \cite{8i}, and Semantic3D.net \cite{semantic3D} datasets. We show the 8x upsampling results based on CD ($10^{-2}$) and PSNR (dB) evaluation metrics. $\uparrow$ ($\downarrow$) means larger (smaller) values correspond to better performance.}
\vspace{-5pt}
\label{main-results}
\begin{center}
\begin{small}
\resizebox{\linewidth}{!}{%
\begin{tabular}{l|cc|cc|cc}
\hline
  & \multicolumn{2}{c|}{ ShapeNet \cite{shapenet}} & \multicolumn{2}{c|}{ 8iVFB \cite{8i}} & \multicolumn{2}{c}{ Semantic3D \cite{semantic3D}} \\
  \hline 
  & CD $\downarrow$ & PSNR $\uparrow$ & CD $\downarrow$ & PSNR $\uparrow$ & CD $\downarrow$ & PSNR $\uparrow$\\
\hline
MPU \cite{mpu}  & 149.20 & 65.37 & 105.43 &66.83 & 181.56 &  63.64 \\
PU-GAN \cite{pugan}  & 174.58 & 64.88 & 117.66 & 66.19& 203.45 & 61.87  \\
Meta-PU \cite{metapu} & 62.05 & 69.72 & 54.31 &  70.48 & 74.86 & 68.24 \\
\hline
PU-GCN \cite{pugcn} & 65.81 & 69.59 & 63.71 & 69.78 &  96.53 & 66.98 \\
\textbf{Ours + PU-GCN} & \textbf{50.49} & \textbf{71.62} & \textbf{41.86} & \textbf{72.18} & \textbf{65.26} & \textbf{69.47} \\
\hline
Dis-PU \cite{dispu} & 55.62 & 70.23 & 51.68 & 70.59 &  67.34 & 69.30 \\
\textbf{Ours + Dis-PU} & \textbf{48.25} & \textbf{71.96} & \textbf{34.65} & \textbf{72.53} & \textbf{54.61} & \textbf{70.18} \\
\hline
PU-Dense \cite{pudense} & 30.52 & 73.11 & 33.18 & 72.57 & 51.25 & 70.62 \\
\textbf{Ours + PU-Dense} & \textbf{26.44} & \textbf{73.38} & \textbf{23.80} & \textbf{73.64} & \textbf{44.63} & \textbf{71.99} \\
\hline
\end{tabular}%
}
\end{small}
\end{center}
\end{table}           

\par We compare the proposed method with five state-of-the-art upsampling methods: MPU \cite{mpu}, PU-GAN \cite{pugan}, PU-GCN \cite{pugcn}, Dis-PU \cite{dispu}, and PU-Dense \cite{pudense}. We also compare our method against a meta-learning-based upsampling approach, namely Meta-PU \cite{metapu}. We apply our framework with several different backbone networks, including PU-GCN \cite{pugcn}, Dis-PU \cite{dispu}, and PU-Dense \cite{pudense}. The comparison is shown in Table \ref{main-results}. All methods are trained on the ShapeNet dataset \cite{shapenet} using the same experimental setup for a fair comparison. As shown in Table \ref{main-results}, our method achieves a significant performance improvement on three backbones \cite{pugcn,dispu,pudense} across all the evaluation metrics with a good margin. More importantly, our method with PU-Dense \cite{pudense} as the backbone outperforms all other state-of-the-art methods. Notably, we observe that the performance improvement on the 8iVFB dataset \cite{8i} and Semantic3D.net dataset \cite{semantic3D} is more significant than the ShapeNet dataset \cite{shapenet}. This demonstrates the effectiveness of our method in boosting the generalization capability of models to unseen test data by enabling the networks to utilize the internal features of point clouds at test time. Besides quantitative results, we present upsampling qualitative comparisons in Figure \ref{fig:qual-results} and Figure \ref{fig:qual-8i-results}. In most cases, the upsampled point clouds of our method are more uniform, less noisy, and preserve edge sharpness.

\subsection{Ablation Studies}\label{sec:ablation}
We perform ablation studies to further analyze our proposed method.

\noindent{\bf Robustness to Noise}: To validate the robustness to noise, we add Gaussian noise of varying noise levels to the input point clouds. We use the model trained on ShapeNet for evaluation
and report our results in Table \ref{noise-results}. As the noise level increases, we can observe that the performance of all approaches drops. But our approach still outperforms all other methods under each noise level by a significant margin. This demonstrates the robustness of our approach to noise.

\begin{table}
\caption{Robustness to upsampling noisy point clouds results with different noise levels on the ShapeNet dataset \cite{shapenet}. We use an upsampling ratio of r=8 and compare different methods using the CD ($10^{-2}$) evaluation metric.}
\vspace{-10pt}

\label{noise-results}
\begin{center}
\begin{small}
\begin{tabular}{l|cccc}
\hline
  & 0\% & 0.5\% & 1\% & 2\%\\
 \hline
PU-GCN \cite{pugcn} & 65.81 & 76.46 & 84.73 & 159.62\\
\textbf{Ours + PU-GCN} & \textbf{50.49} & \textbf{56.32} & \textbf{63.61} &\textbf{124.79}\\
\hline
Dis-PU \cite{dispu} & 55.62 & 65.24 & 71.19 & 136.87 \\
\textbf{Ours + Dis-PU} & \textbf{48.25} & \textbf{52.58} & \textbf{59.35} &\textbf{117.52}\\
\hline
PU-Dense \cite{pudense} & 30.52 & 35.71 & 37.55 & 74.67 \\
\textbf{Ours + PU-Dense} & \textbf{26.44} & \textbf{29.10} & \textbf{33.96} & \textbf{62.41}\\
\hline
\end{tabular}
\end{small}
\end{center}
\end{table}

\begin{table}
\caption{Quantitative comparisons with baselines on the ShapeNet dataset \cite{shapenet} with varying upsampling scale ratios.}
\vspace{-10pt}
\label{ratio-results}
\begin{center}
\begin{small}
\begin{tabular}{l|cc|cc}
\hline
  & \multicolumn{2}{c}{ 4x } & \multicolumn{2}{c}{ 16x}\\
  \hline 
  & CD $\downarrow$ & PSNR $\uparrow$ & CD $\downarrow$ & PSNR $\uparrow$\\
\hline
PU-GCN \cite{pugcn} & 48.15& 70.90 & 121.65 & 66.27 \\
\textbf{Ours + PU-GCN } & \textbf{31.74} & \textbf{72.87} & \textbf{102.84} & \textbf{67.11}\\
\hline
Dis-PU \cite{dispu} & 36.23& 72.19 &  92.88 & 67.46\\
\textbf{Ours + Dis-PU} & \textbf{25.61} & \textbf{73.28} &  \textbf{84.31}& \textbf{69.01}\\
\hline
PU-Dense \cite{pudense} & 18.82 & 75.24 & 69.48 & 70.32\\
\textbf{Ours + PU-Dense} & \textbf{15.33} & \textbf{75.86} & \textbf{62.19} & \textbf{70.73} \\
\hline
\end{tabular}
\end{small}
\end{center}
\end{table}

\noindent{\bf Varying Upsampling Ratios}: We further investigate the robustness of our framework across different upsampling ratios. We have conducted experiments with upsampling scale ratios r = 4, 8, and 16. Table \ref{main-results} reports the results with upsampling scale ratio of 8x on ShapeNet \cite{shapenet} and 8iVFB \cite{8i} datasets. Table \ref{ratio-results} shows the quantitative comparisons on the ShapeNet \cite{shapenet} dataset under upsampling scales of 4x and 16x. We can observe that our method effectively improves the performance of all backbones across different upsampling ratios.

\begin{table}
\caption{Ablation studies on different components of our framework components, including meta-learning and test-time adaptation.}
\vspace{-10pt}

\label{ablation1}
\begin{center}
\begin{small}
\begin{tabular}{l|cc}
\hline
  & CD $\downarrow$ & PSNR $\uparrow$ \\
 \hline
PU-GCN \cite{pugcn} & 65.81 & 69.59  \\
PU-GCN + TTA (w/o meta) & 61.74 & 69.83 \\
\textbf{Ours + PU-GCN} & \textbf{50.49} & \textbf{71.62} \\
\hline
Dis-PU \cite{dispu} & 55.62 & 70.23\\
Dis-PU + TTA (w/o meta)  & 54.52 & 70.40\\
\textbf{Ours + Dis-PU} & \textbf{48.25} & \textbf{71.96} \\
\hline
PU-Dense \cite{pudense} & 30.52 & 73.11 \\
PU-Dense + TTA (w/o meta) & 28.33 & 73.18 \\
\textbf{Ours + PU-Dense} & \textbf{26.44} & \textbf{73.38} \\
\hline
\end{tabular}
\end{small}
\end{center}
\end{table}
\begin{table}
\caption{Ablation studies on the number of gradient updates ($N = 1, 3, 5, 7, 9$). We report the 8x results of the ShapeNet dataset \cite{shapenet} based on CD ($10^{-2}$), PSNR (dB), and inference time (ms).}
\vspace{-10pt}

\label{gradient-results}
\begin{center}
\begin{small}
\begin{tabular}{l|ccc}
\hline
  & CD $\downarrow$ & PSNR $\uparrow$ & Time\\
 \hline
Dis-PU \cite{dispu} & 55.62 & 70.23 & \textbf{36.31} \\
Ours + Dis-PU (N=1) & 53.41 & 70.86 & 50.46 \\
Ours + Dis-PU  (N=3)  & 50.71 & 71.89 &  95.38 \\
Ours + Dis-PU (N=5) & 48.25 & \textbf{71.96} & 148.65 \\
Ours + Dis-PU (N=7) & \textbf{47.83} & 71.95 & 197.28 \\
Ours + Dis-PU (N=9) & 48.79 & 71.92 & 254.72 \\
\hline
\end{tabular}
\end{small}
\end{center}
\end{table}
\noindent{\bf Framework Components}: To study the relative contributions of various components in the proposed framework, we conduct additional ablation experiments. We first investigate the effect of test-time adaptation on improving performance. In this experiment, we do not apply our meta-learning approach. Instead, we use the pre-trained parameters resulting from supervised training. At test time, the input point cloud is downsampled and the model is fine-tuned using the input and the downsampled point clouds. As shown in Table \ref{ablation1}, the performance of all the backbones has already been improved. This demonstrates the effectiveness of naive test-time adaptation in utilizing the internal features of point clouds, even without meta-learning. When applying our meta-training approach in Algorithm \ref{alg:algoTrain}, the performance has been further improved. This demonstrates that both TTA and meta-training contribute to the final performance improvement.

\noindent{\bf Number of Gradient Updates}: In this study, we investigate the impact of the number of gradient updates $N$ in the inner loop of Algorithm \ref{alg:algoTrain}. We use N = 1, 3, 5, 7, and 9 during the meta-training. Table \ref{gradient-results} shows the evaluation results of our method trained with a different number of gradient updates. Overall, we observe that a large number of gradient updates enables the model to better capture the internal features of test point clouds and thus improve the performance. However, the PSNR with N=7 is slightly
worse when compared to N=5. Furthermore, the performance experiences a minor decrease with N=9. Note that, we use the same number of gradient updates during training and testing. 

 \section{Conclusion}
In this paper, we have introduced a novel test-time adaptation framework for point cloud upsampling that utilizes both internal and external features of point clouds. In previous work, the model is typically trained on an external supervised dataset and fixed during evaluation on unseen test data. This approach fails to exploit the useful internal information of the test point clouds. In contrast, our framework is designed to efficiently adapt the model parameters for each test instance at inference time to improve the upsampling performance. To this end, we have proposed a meta-learning algorithm that allows fast adaptation of model parameters at test time using only the input sparse point cloud. More importantly, our method is a generic framework that can be applied to any deep learning-based upsampling network without modifying the architecture. Extensive experiments show the effectiveness of our proposed approach in boosting the upsampling performance and outperforming state-of-the-art methods.

{{\small
\bibliographystyle{IEEEtran}
\bibliography{iros}
}}

\end{document}